\documentclass[runningheads]{llncs}
\usepackage{graphicx}

\usepackage{tikz}
\usepackage{comment}
\usepackage{amsmath,amssymb} 
\usepackage{color}
 
\usepackage[accsupp]{axessibility}

\usepackage[misc,geometry]{ifsym}

\usepackage{enumitem}
\usepackage{epsfig}
\usepackage{amsmath}
\usepackage{amssymb}
\usepackage{bm}
\usepackage{graphicx}
\graphicspath{{figs/}}
\usepackage{url}            
\usepackage{booktabs}       
\usepackage[utf8]{inputenc} 
\usepackage[T1]{fontenc}    
\usepackage{amsfonts}       
\usepackage{nicefrac}       
\usepackage{microtype}      
\usepackage{xcolor}
\usepackage{pifont}
\usepackage{wasysym}
\usepackage{amssymb}
\usepackage{multirow}
\usepackage{subfigure}
\usepackage{caption}
\usepackage{rotating}
\usepackage{cite}
\usepackage{array}
\usepackage{bm}
\usepackage{color}
\usepackage{pifont}
\usepackage{diagbox}
\newcommand{\sArt}{state-of-the-art }
\def\ie{\emph{i.e.}}
\def\eg{\emph{e.g.}}
\def\etc{\emph{etc}}
\def\etal{{\em et al.}}

\begin{document}
 
\pagestyle{headings}
\mainmatter
\def\ECCVSubNumber{5613}  

\title{Salient Object Detection for Point Clouds} 

\titlerunning{Salient Object Detection for Point Clouds}
 
\author{Songlin Fan\inst{1,2} \and
	Wei Gao\inst{1,2,}\textsuperscript{\Letter} \and
	Ge Li\inst{1}}
 
\authorrunning{Songlin Fan \etal}
 
\institute{Peking University Shenzhen Graduate School \and
	Peng Cheng Laboratory\\
	\email{\{slfan, gaowei262, geli\}@pku.edu.cn}\\
	\textcolor{blue}{\url{https://git.openi.org.cn/OpenPointCloud/PCSOD}}}

\maketitle

\begin{abstract}
	This paper researches the unexplored task---point cloud salient object detection (SOD). Differing from SOD for images, we find the attention shift of point clouds may provoke saliency conflict, \ie, an object paradoxically belongs to salient and non-salient categories. To eschew this issue, we present a novel view-dependent perspective of salient objects, reasonably reflecting the most eye-catching objects in point cloud scenarios. Following this formulation, we introduce \textbf{\textit{PCSOD}}, the first dataset proposed for point cloud SOD consisting of 2,872 in-/out-door 3D views. The samples in our dataset are labeled with hierarchical annotations, \eg, super-/sub-class, bounding box, and segmentation map, which endows the brilliant generalizability and broad applicability of our dataset verifying various conjectures. To evidence the feasibility of our solution, we further contribute a baseline model and benchmark five representative models for a comprehensive comparison. The proposed model can effectively analyze irregular and unordered points for detecting salient objects. Thanks to incorporating the task-tailored designs, our method shows visible superiority over other baselines, producing more satisfactory results. Extensive experiments and discussions reveal the promising potential of this research field, paving the way for further study.
	\begin{figure*}[t]
		\centering
		\includegraphics[width=0.85\linewidth]{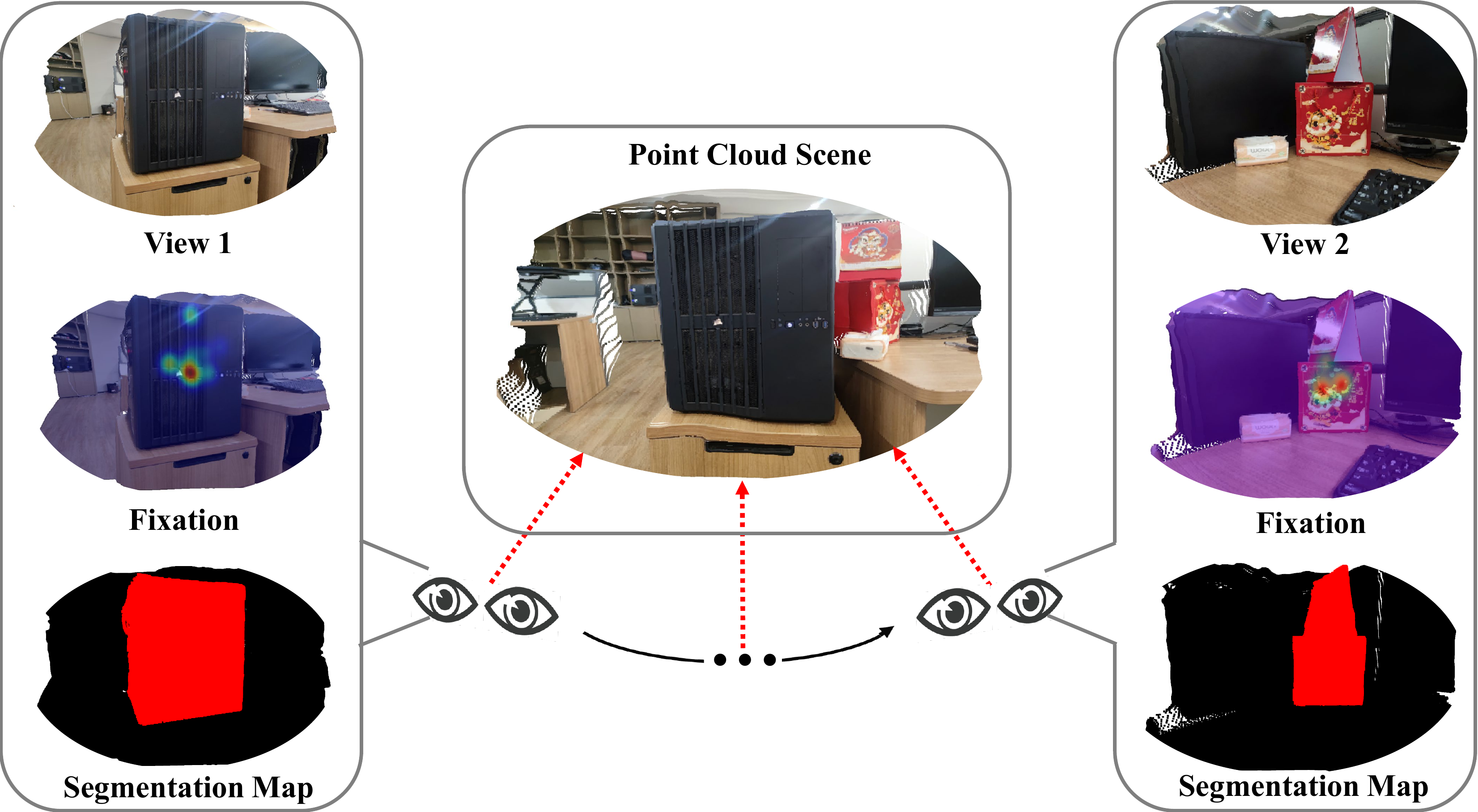}
		\caption{Illustration of saliency conflict. The variation of attention allocated to the black computer causes a contradiction that the black computer simultaneously belongs to the salient and non-salient objects for this scene. We, therefore, propose to analyze the salient objects of point clouds according to the views.}
		\label{fig1}
	\end{figure*}
	
\end{abstract}

\section{Introduction}

Salient objects describe the most attractive objects with respect to their surroundings. Due to its myriad applications, salient object detection (SOD) can provide the pre-processing results for many vision tasks, such as 3D shape classification~\cite{7410384},  compression~\cite{9522973}, and quality assessment~\cite{9756929}, to name a few. Distinct from the relevant task~\cite{8726371,6751558,9010640} for predicting eye fixation positions, namely saliency detection, SOD demands locating salient objects and completely segmenting them further, thus being more challenging. Most existing SOD works~\cite{8315520,9439490,zhang2020asymmetric,8954050,liao2020mmnet,piao2020exploit} devote their efforts to analyzing salient objects on regular images. With the fast revolution of 3D collection equipment, point clouds as the raw output of many devices (such as LiDAR and depth sensors) have a growing presence in research and applications. Compared with the adoption of alternative 3D formats, data processing directly on point clouds avoids information loss and computational redundancy in format conversion that may induce performance drops. Despite the flourishing advance of many point-based tasks, \eg, classification~\cite{8099499}, object detection~\cite{8954080}, and segmentation~\cite{9440696}, point cloud SOD is still in its infancy, and many issues have not been discussed yet.

As immersive visual media, point clouds offer a watching experience with six degrees of freedom (6DOF). Unlike the watching of static images, the attention allocation of humans varies when the view changes. The research community dubs the phenomenon that attention being allocated from one region to another as the attention shift~\cite{9156279,8954050}. However, we find that the attention shift of point clouds may trigger a new thorny problem that we name {saliency conflict}, \ie, an object paradoxically belongs to salient and non-salient categories for different views of one point cloud scene sample. Fig.~\ref{fig1} shows an example of an office scene recorded by point clouds. The attention allocated to the black computer varies as the view changes, which causes the black computer to go from being the salient object to the non-salient object. \textbf{Then is the black computer the salient object of this office scene?} The answer matters not only the definition of salient objects in point cloud scenarios but also the relevant dataset construction.

In this paper, we argue that the manifestations of salient objects in point clouds depend on the views, and point cloud SOD is to compute the salient objects of any given view in 3D space. The union of salient objects (segmentation maps) of “given views” indicates the complete description of salient objects for scenes in point clouds. For Fig.~\ref{fig1}, different segmentation maps correspond to different views, and the union of segmentation maps represents the salient objects of this office scene. Firstly, this formulation makes it easier to grasp the nature of the SOD problem due to the fact that humans actually observe only one view at a time while the viewpoint is free. Broadly speaking, the image is a special case of only a single view. Secondly, this formulation avoids the complex modeling to handle the whole 3D scene with saliency conflict phenomenon, which can benefit the design of simpler models capable of analyzing different views. Thirdly, this formulation eases the dataset construction with the human-annotated most attractive objects via subjective experiments, since the subjective experiment results of different views sometimes cannot be reflected into a large-scale point cloud sample (such as the office scene sample in Fig.~\ref{fig1}) simultaneously without our view-dependent saliency analysis.

Following our formulation of point cloud SOD, we introduce \textit{PCSOD}---the first versatile dataset for point cloud SOD with densely annotated labels. Our dataset contains 2,872 frequent 3D views that belong to over one hundred in-/out-door scenes. The manual data collection phase lasts over one year, and the samples reflect a wide range of scenarios in our lives. Detailed statistics show that our dataset has 138 object categories and 53.4\% difficult samples, which ensures its brilliant generalizability. To extend the applicability of this new dataset, we provide hierarchical annotations for each sample, including super-/sub-class, bounding box, and segmentation map. The proposed dataset as a comprehensive platform can conveniently support research on multi-task learning~\cite{9336293} and other valuable vision tasks, not limited to point cloud SOD. 

Since point clouds record 3D information in the format of irregular and unordered points, existing SOD models~\cite{8315520,9439490,zhang2020asymmetric,8954050,liao2020mmnet} for images cannot be transferred for point cloud processing. Additionally, though several representative point-based models~\cite{8099499,qi2017pointnet++,li2018pointcnn,9440696,9010996} have been developed for other segmentation tasks, they are incapable of performing well in SOD. These models for other segmentation tasks fail to consider the particularities of SOD, \ie, the benefits of multi-scale features~\cite{9156530} and the refinement of global semantics~\cite{chen2020global,liu2019simple}. To prove the feasibility of our solution, we further develop a baseline model and benchmark five representative segmentation models for comparison and analysis of point cloud SOD. Owing to incorporating the task-tailored designs, the proposed baseline model can take full advantage of the multi-scale features and global semantics to locate salient objects and accurately separate them. Extensive experiments verify the effectiveness of our solution for point cloud SOD.

In summary, we conclude the contributions as follows:
\begin{itemize}
	\item[1)] We propose a novel view-dependent perspective of point cloud SOD. Our formulation avoids the {saliency conflict}, emphasizes the nature of SOD, and reasonably reflects the most eye-catching objects in point cloud scenarios.
	\item[2)] We construct the first versatile dataset for point cloud SOD, termed \textit{{PCSOD}}. Our dataset has brilliant generalizability and broad applicability, expected to be a catalyst for point cloud SOD and many other vision tasks.
	\item[3)] We develop a baseline model for point cloud SOD. Our baseline model has a full consideration of the particularities of SOD, outperforming other baseline models by a clear margin.
	\item[4)] We establish the first benchmark of point cloud SOD, conduct a thorough analysis, and bring a new perspective toward point cloud SOD.
\end{itemize}

\section{Related Work}
\subsubsection{Salient Object Detection.}
Following the pioneer attempt~\cite{730558}, many early works~\cite{wei2012,7780625,6247743,6751481} design hand-crafted features to exploit low-level cues. These methods cannot obtain satisfactory accuracy because of the lack of semantic cues. Thanks to the powerful capability of neural networks in abstracting semantics, the bottleneck of traditional methods is broken. Hou~\etal~\cite{8315520} introduce short connections into a skip-layer structure. The advanced representations at multiple layers thus can be fully utilized. Siris~\etal~\cite{siris2021scene} propose a semantic scene context-aware framework to capture sufficient high-level semantics for locating salient objects. To rich the semantic information diluted during the top-down transmission, some recent works~\cite{chen2020global,liu2019simple, 9156530} explicitly extract global semantics and append them into low-level features, achieving visible performance improvement. Despite the gratifying achievements of existing RGB image-based methods~\cite{9557833,9008371,8953756,7780449}, they still have difficulty understanding complex scenes for lacking spatial geometry information. Consequently, researchers begin extending the task of SOD on 3D images, such as RGB-D images~\cite{lang2012depth,8315520,zhang2020asymmetric,liao2020mmnet,li2020rgb,9156838,fan2020bbs} and light field images ~\cite{7570181,8984718,9010926,liu2021light}, which show significant potential. A detailed description of these image-based methods is beyond the scope of this article. Please refer to the relevant surveys~\cite{9320524,zhou2021rgb,fu2020light} for more introduction.  We can conclude that all these efforts are confined to the image domain. This work will disentangle the limitation and probe SOD on point clouds.

Regarding the attention modeling on point clouds, we also learn that a few methods~\cite{kim2008segmentation,guo2018point,8726371,6751558,9010640,7410384} are developed to automatically compute the human attention distribution.  The algorithms of these methods merely produce a heatmap of the attention distribution, while the SOD task we study demands completely segmenting the salient objects, thus being more challenging.

\subsubsection{Deep Learning on Point Clouds.}
Processing point clouds has long been a significant challenge. Previous works~\cite{8237361,8953410} tend to first rearrange raw points via octree or kdtree. The emergence of PointNet/PointNet$++$~\cite{8099499,qi2017pointnet++} shows us a new approach for raw point processing. They employ shared multilayer perceptrons (MLPs) to extract point-wise features and achieve \sArt performance across many vision tasks. Following PointNet, three directions are mainly adopted to improve the performance further, \ie, powerful convolution~\cite{li2018pointcnn,8954200}, effective  neighborhood connection~\cite{9010996,wang2019dynamic}, and advanced reduction~\cite{9440696, qian2021assanet}.   Li~\etal~\cite{li2018pointcnn} propose to learn an X-transformation from raw points by imitating the typical convolution, while Wu~\etal~\cite{8954200} regard the typical convolution as the combination of weight and density functions. ShellNet~\cite{9010996} arranges neighbors into concentric spherical shells that have a convolution order from the inner to the outer shells. Wang~\etal~\cite{wang2019dynamic} propose a simple operation known as EdgeConv, which extracts local geometric features while retaining permutation invariance. To explore more advanced reduction operations, Hu~\etal~\cite{9440696} and Qian~\etal~\cite{qian2021assanet} resort to attentive pooling and anisotropic reduction, respectively. However, these methods are not initially developed for SOD, ignoring the particularities of SOD.

\begin{figure*}[t]
	\centering
	\includegraphics[width=\linewidth]{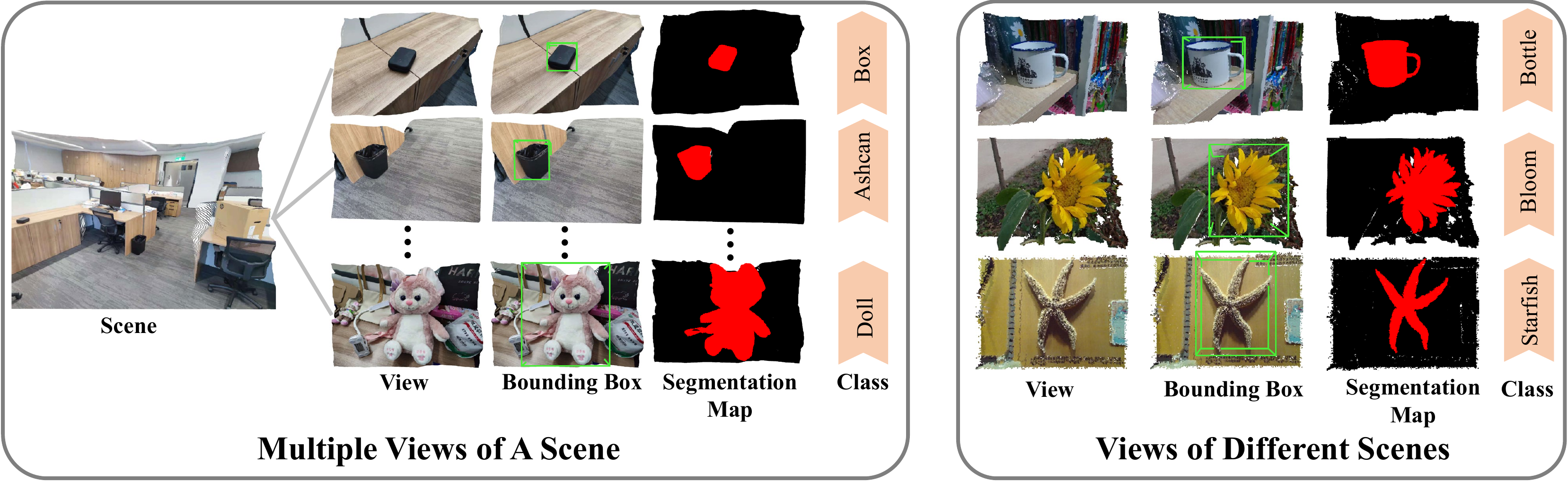}
	\caption{Examples from our \textit{PCSOD} dataset with hierarchical annotations.}
	\label{fig2}
\end{figure*}

\section{Proposed Dataset}
Datasets~\cite{russakovsky2015imagenet,7570181,8954050} have become the driving force behind many vision tasks, especially with the emergence of deep learning. With this in mind, we introduce \textit{PCSOD} for: (1) probing a new challenging task, (2) facilitating research on new issues, and (3) verifying new conjectures. Next, we will elaborate more details about our dataset. Besides, some visual examples are shown in Fig.~\ref{fig2} and Fig.~\ref{fig3}.

\subsection{Dataset Construction}  

\subsubsection{Data Collection.} Point clouds in existing datasets~\cite{behley2019semantickitti,armeni20163d,hackel2017semantic3d,dai2017scannet,mo2019partnet} are often collected for specific scenes (such as outdoor road or indoor office scenes). In contrast, a high-quality SOD dataset~\cite{8099887} demands rich scenes, which motivates us to collect diverse data by ourselves. The data collection phase takes over one year, and we collect 2,872 3D views from over one hundred preset scenes across dozens of cities. Each 3D view has 240,000 points. This process can also simulate the 3D view acquisition when “travelling” in an off-shelf large-scale point cloud sample (such as an office or even a city). As shown in Fig.~\ref{fig2}, the 3D views of a scene constitute a series of watching descriptions of this scene whose salient objects can be obtained from subjective experiments without saliency conflict. 

\begin{figure*}[t]
	\centering
	\includegraphics[width=\linewidth]{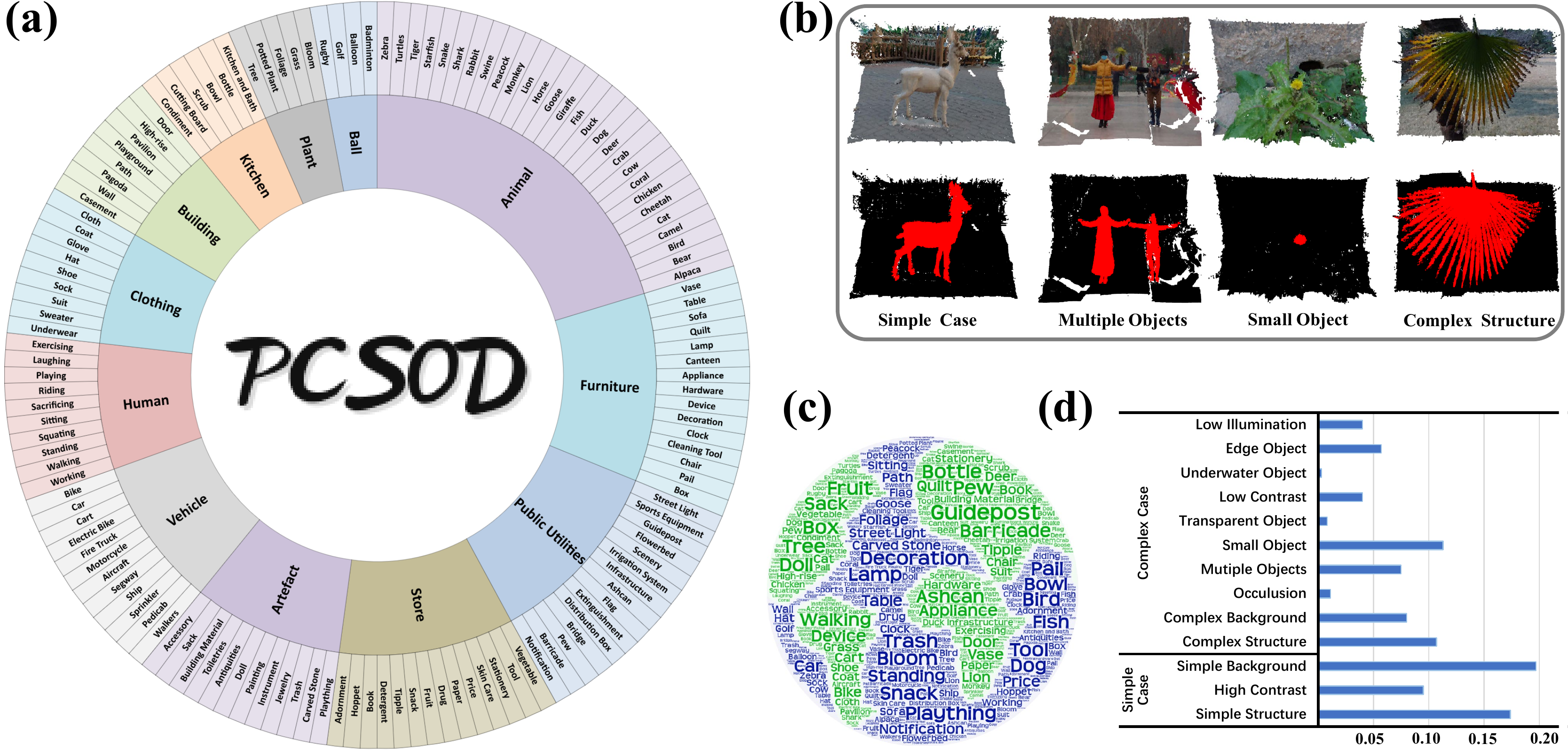}
	\caption{Statistics of our \textit{PCSOD} dataset. (a) Categories of salient objects. (b) Illustration of challenging samples.  (c) Word cloud of salient objects. (d) Histogram distribution of challenging samples.}
	\label{fig3}
\end{figure*}

\subsubsection{Data Annotation.} Referring to the determination of salient objects in images~\cite{9010926, 8099887}, we employ thirty professional annotators to label the salient objects from given views. Before the labeling, every annotator is pre-trained over fifteen samples. To ensure the annotation accuracy, we divide these thirty annotators into ten groups. Three annotators in one group jointly determine the salient objects, then cross-validated by other groups. An object is regarded as a positive label only if more than eighty percent of annotators verify it. The recently released datasets~\cite{mo2019partnet,8954050} indicate that offering hierarchical annotations benefits the applicability of a new dataset. As shown in Fig.~\ref{fig2}, we hierarchically label the determined salient objects and provide three levels of annotations, \ie, class, bounding box, and segmentation map. Each level of annotations is obtained through corresponding professionals. Furthermore, at least two passes of verification are performed for each annotation to ensure its quality. 

\subsubsection{Data Split.} Having a standard dataset split~\cite{8954050,9010926} is conducive to fairly studying and comparing the pros and cons of algorithms. Following the ratio of 7:3 adopted by many datasets~\cite{9010926}, our \textit{PCSOD} is randomly split into 2,000 samples for training and 872 samples for testing. 

\subsection{Dataset Statistics}

\subsubsection{Diverse Object Categories.} A diverse SOD dataset should have broad coverage of scenes in the real world to ensure brilliant generalizability. Our \textit{PCSOD} covers a wide range of scenarios in our lives. As shown in Fig.~\ref{fig3}(a) and Fig.~\ref{fig3}(c), the salient object categories have a heterogeneous variety. Specifically, objects in our dataset can be categorized into 12 super-classes, \eg, human, animal, plant, \etc. These 12 super-classes are further comprised of 138 sub-classes,  fully covering the daily situations. The diverse salient object categories enable a comprehensive understanding of the attention allocation of humans in real-world scenes.

\subsubsection{Rich Annotations.} A versatile dataset should not only support the study of existing issues but also adapt to new research directions. As shown in Fig.~\ref{fig2}, our \textit{PCSOD} offers hierarchical
annotations, \eg, super-/sub-class, bounding box, and segmentation map. These annotations help researchers understand each sample of our dataset from different aspects (such as object property, object proposal, and scene parsing), sparking novel ideas. Besides, our annotations are very precise. The segmentation maps accurately reflect the structures of objects in 3D scenes, even though some are very complex (see the complex structure case in Fig.~\ref{fig3}(b)).

\subsubsection{Difficult Samples.}
A valuable dataset should contain a certain amount of difficult samples and dive into the problems. The difficult samples benefit the performance of models confronting various complex scenes.  With this consideration, we add many challenging samples to our dataset, including multiple objects, small objects, complex structures, low illumination, \etc. Some visual examples are shown in Fig.~\ref{fig3}(b). Fig.~\ref{fig3}(d) further details the proportion of samples with each attribute. Statistics indicate that our dataset has 53.4\% difficult samples, which evidences that the proposed \textit{PCSOD} is very challenging.

\section{Proposed Method}

Extending the concept of salient objects in images to point clouds, we formulate that the salient objects of views from a scene indicate the complete description of salient objects in this scene. Point cloud SOD aims to identify the salient objects of any given view. While various methods have been developed for image-based SOD, they cannot handle irregular and unordered point clouds. Moreover, existing point-based segmentation models for other tasks cannot guarantee the performance of identifying salient objects. These circumstances motivate us to design a baseline model and excavate potential directions for point cloud SOD.

\subsection{Overall Architecture}
\begin{figure*}[t]
	\centering
	\includegraphics[width=\linewidth]{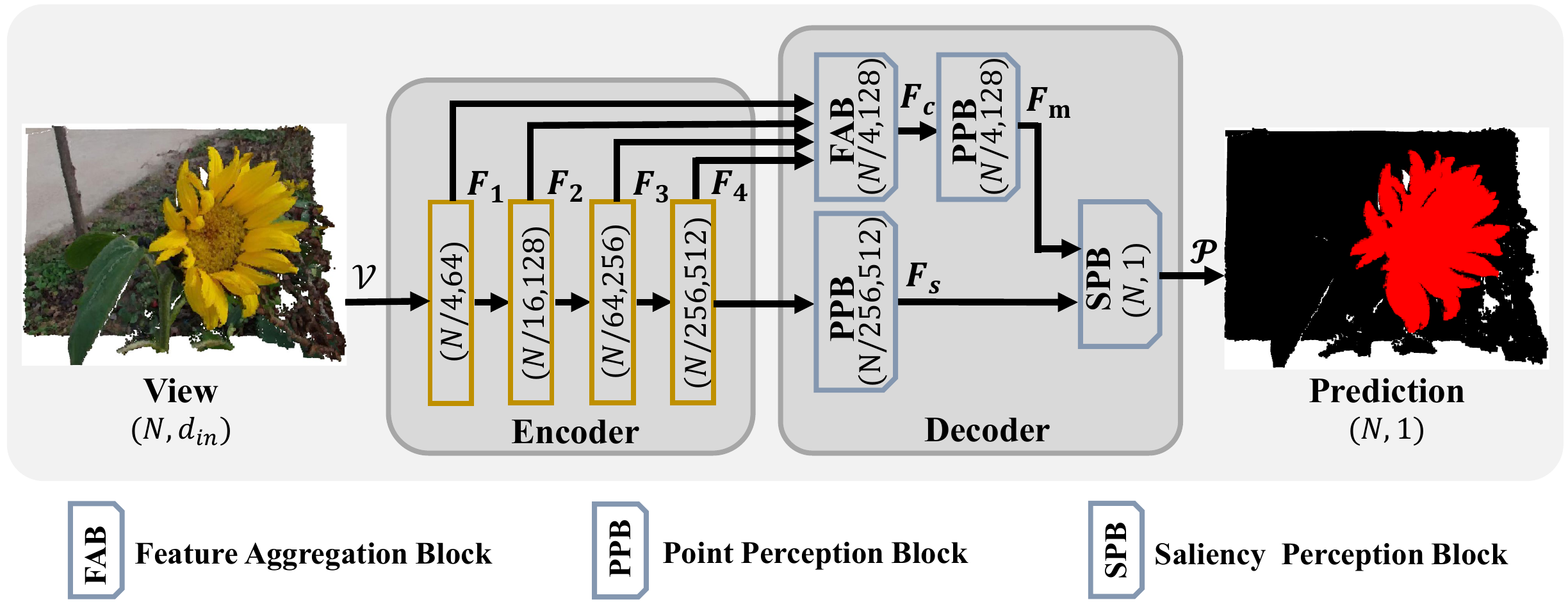}
	\caption{Overall architecture of the proposed baseline model, which has a typical encoder-decoder architecture.}
	\label{fig4}
\end{figure*}

As shown in Fig.~\ref{fig4}, the proposed baseline model inherits a typical encoder-decoder architecture. The encoder extracts multi-level features from raw points, while the decoder enhances and fuses the extracted features to predict salient objects. To illustrate the effectiveness of our designs, we introduce the classical PointNet$++$~\cite{qi2017pointnet++} as the encoder. It has been studied~\cite{liu2019simple} that high-level features will be gradually diluted when transmitted to low-level ones. To address this issue, some recent image-based methods~\cite{chen2020global,liu2019simple, 9156530} explicitly extract global semantics and append them into low-level features, observing gratifying performance improvement. Inspired by the philosophical designs of these methods, we design two key modules, \ie, Point Perception Block (PPB) and Saliency Perception Block (SPB), to take full advantage of the benefits of multi-scale features and the refinement of global semantics for locating salient objects.

Formally, let $\mathcal{V}=\{v_1,v_2,...,v_{N}\}$ represent a view of $N$ points with associated point-wise features (\eg, RGB colors), where $v\in\mathbb{R}^{d_{in}}$. To obtain the probabilities $\mathcal{P}=\{p_1,p_2,...,p_N\}$ of corresponding points being salient, the encoder first extracts multi-level features $\{F_l\}_{l=1}^4$ from raw points $\mathcal{V}$. The $l^{th}$ level features $F_l=\{f^{l}_{1},f^{l}_{2},...,f^{l}_{N_l}\}$ have $N_l=\frac{N}{4^l}$ aggregated points with doubling the feature dimension compared with $F_{l-1}$ (except the feature dimension of the first level is fixed to 64).  Then we aggregate multi-level features $\{F_l\}_{l=1}^4$ into the compact representations $F_c$ via the Feature Aggregation Block (FAB). As shown in Fig.~\ref{fig5}, the operations in FAB are very straightforward, \ie, upsampled high-level features are sequentially fused with low-level features. We adopt the common trilinear interpolation as the upsampling operation to match the spatial size of different level features, while the fusion operation we employ is concatenation along the feature dimension followed by MLPs. Following previous works~\cite{qi2017pointnet++,9440696}, the feature concatenation can simultaneously retain the originality of the fused two level features and is proved to be very effective for point cloud feature fusion. Note that the feature fusion in all modules is uniformly through concatenation unless otherwise stated. To prevent the dilution of high-level features, the PPB is proposed to abstract global semantics and strengthen the multi-scale representations. We obtain global semantics $F_s$ and multi-scale features $F_m$ from the highest-level features $F_4$ and the compact representations $F_c$, respectively, using two PPBs with different configurations. The global semantics can supplement the diluted high-level features in multi-scale features and alleviate the distraction of non-salient background. To achieve this, we further develop the SPB to integrate multi-scale features $F_m$ and global semantics $F_s$, and produce the final prediction $\mathcal{P}$. Next, we will elaborate on the details of our PPB and SPB.

\subsection{Proposed Modules}

\subsubsection{Point Perception Block.}
The global semantics and multi-scale features are important for SOD~\cite{chen2020global,liu2019simple, 9156530}. The former helps to locate the positions of salient objects, while the latter is conducive to recognizing salient objects of different sizes. Besides, the acquisition of them demands enlarging the receptive fields of features and capturing the context information. Inspired by the widely used Receptive Field Block~\cite{liu2018receptive}, we introduce the PPB to achieve this goal. 

\begin{figure*}[t]
	\centering
	\includegraphics[width=\linewidth]{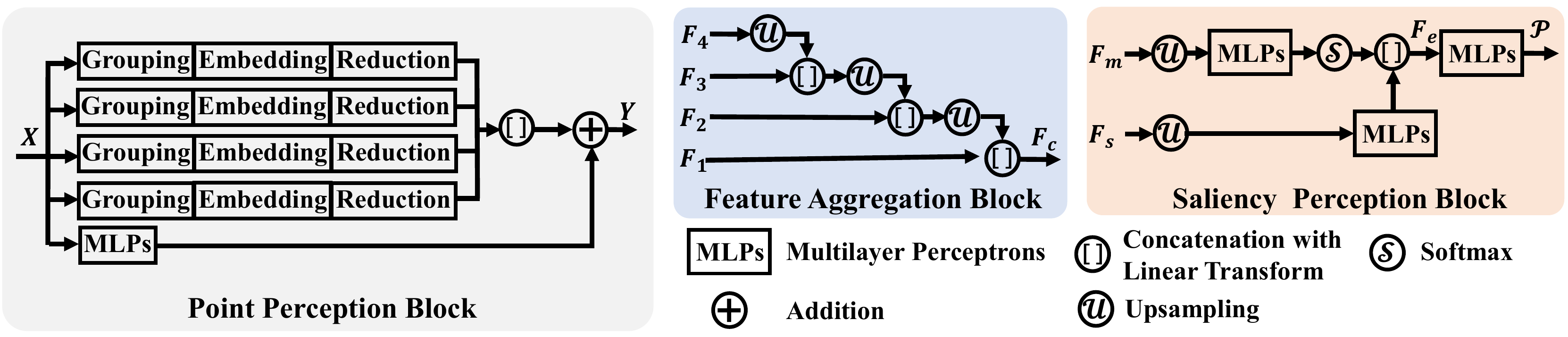}
	\caption{Details of the components in the proposed baseline model, \ie, Point Perception Block, Feature Aggregation Block, and Saliency Perception Block.}
	\label{fig5}
\end{figure*}

As shown in Fig.~\ref{fig5}, the PPB consists of five branches to capture the context information of point-wise features. The first four branches with similar structures encode  center points by their local regions of different sizes. Each branch has three sub-units, \ie, grouping, embedding, and reduction. To be more specific, let $X^p=\{x^P_1,x^P_2,...,x^p_M\}$ denote the spatial coordinates of input points $X$ with intermediate learned features $X^f=\{x^f_1,x^f_2,...,x^f_M\}$. $M$ indicates the number of points. For each point $x^p_i\in X^p$, the grouping sub-unit gathers its $k$ nearest neighbors $\mathcal{N}(x^p_i)=\{x^p_{i,1}, x^p_{i,2},..., x^p_{i,k}\}$ by K-nearest
neighbours (KNN). The spatial size of the local region $\mathcal{N}(x^p_i)$ centered on $x^p_i$ varies as $k$ takes different values. To learn local geometric representations, the embedding sub-unit embeds the relative spatial position between $x^p_i$ and its neighbor $x^p_{i,j}$  as 
\begin{equation}
\label{eq1}
e_i^j=MLPs([x^p_i,x^p_{i,j},x^p_i-x^p_{i,j},\mathcal{D}(x^p_i,x^p_{i,j})])\text{,}
\end{equation}
where $\mathcal{D}(\cdot)$ and $[]$ denote the Euclidean distance between two points and the concatenation operation, respectively. Because $e_i^j$ merely contains the geometric features and lacks  associated point-wise features, we concatenate $e_i^j$ with corresponding point-wise features $x^f_{i}$ to obtain the advanced representations $a_i^j$. All advanced representations $\mathcal{A}_i=\{a_i^1,a_i^2,..,a_i^k\}$ of $k$ neighbors express each of their semantic contributions to the center point $x^p_i$. The reduction sub-unit aggregates the neighborhood semantic contributions by a Mean-max reduction operation
\begin{equation}
\label{eq2}
\hat{x}_i^f=MLPs([max(\mathcal{A}_i),mean(\mathcal{A}_i)])\text{,}
\end{equation}
where $max(\cdot)$ and $mean(\cdot)$ denote the max function and mean function, respectively. Compared with the input features $X^f$, the branch outputs $\hat{X}^f=\{\hat{x}^f_1,\hat{x}^f_2,...,\hat{x}^f_M\}$ have enlarged receptive fields and capture the context information in local regions.  Finally, we fuse the output features $\{\hat{X}^f_b\}^4_{b=1}$ of the first four branches and further introduce a skip connection of the fifth branch to retain the original features
\begin{equation}
\label{eq3}
\hat{Y}^f=MLPs([\hat{X}^f_1,\hat{X}^f_2,\hat{X}^f_3,\hat{X}^f_4]) + MLPs(X^f) \text{.}
\end{equation}

Similar to the Receptive Field Block, by setting $K=\{k_1,k_2,k_3,k_4\}$ for the first four branches reasonably, the global semantics and multi-scale features can be obtained, respectively. Besides, the input points $X$ and corresponding outputs $Y$ of our PPB share the same feature size. Therefore, our PPB can be easily embedded in various networks to improve their performance.

\subsubsection{Saliency Perception Block.} The utilization of our PPB allows the acquisition of global semantics and multi-scale features. Subsequently, how to seamlessly merge the two kinds of features and obtain the final prediction is still open.

As shown in Fig.~\ref{fig5}, our SPB enhances the multi-scale features using the global semantics. The global semantics can effectively alleviate the distraction of non-salient background in multi-scale features and emphasize the salient regions (see Fig.~\ref{fig8}). The enhanced multi-scale features are then used to predict the salient objects. Specifically, the SPB first upsamples the global semantics $F_s$ and multi-scale features $F_m$ to the spatial size of the input $\mathcal{V}$. The upsampling operation is followed by MLPs to reduce the aliasing effect. Then we use the upsampled global semantics to enhance the multi-scale features
\begin{equation}
\label{eq4}
F_e = MLPs([MLPs(\mathcal{U}(F_s)),\mathcal{S}(MLPs(\mathcal{U}(F_m)))])\text{,}
\end{equation}
where $\mathcal{U}$ and $\mathcal{S}$ denote the upsampling and softmax operations, respectively. $F_e$ is the enhanced multi-scale features. In this approach, the enhanced multi-scale features include both the accurate positions and fine-grained structures of salient objects. Finally, we use a prediction layer (MLPs) to predict salient objects $\mathcal{P}$ from the enhanced multi-scale features $F_e$.

\section{Experiments}
\label{exp}
\subsection{Experimental Setup}
\subsubsection{Implementation Details.} We use the popular Pytorch framework to implement our method on an NVIDIA Tesla V100 GPU. The points in the inputs are represented by nine-dimensional vectors ($d_{in}=9$) consisting of spatial coordinates, RGB colors, and normalized spatial coordinates.  Due to the limitations of memory capacity, we randomly sample $N=4,096$ points with replacement from inputs in the training stage, while the sampling operations in the testing are without replacement for testing all 240,000 points in a 3D view. We use random rotation to augment data. The parameters $K$ of the PPB for abstracting global semantics are $\{1,4,9,16\}$ while those of another PPB are $\{1,9,25,49\}$. Our loss function is defined on the standard cross-entropy loss. We train the proposed baseline model by Adam optimizer with an initial learning rate of 5e-4 and a weight decay of 1e-4. The total training epochs are 3,000, with a batch size of 32. A three-time voting strategy~\cite{qi2017pointnet++} is adopted to produce the predictions in the testing phase.

\subsubsection{Evaluation Metrics.} To compare the results of different methods, we adopt four popular evaluation metrics for performance benchmarking, \ie, mean absolute error (MAE), F-measure~\cite{6909433}, E-measure~\cite{fan2018enhanced}, and intersection over union (IoU). MAE estimates the point-wise approximation degree between predicted segmentation maps and corresponding ground truths. It can be formulated as $ \text{MAE}=\frac{1}{N}\sum^N_{i=1}|p_i-g_i|$, where $p_i \in \mathcal{P}$ and  $g_i \in \mathcal{G}$ are the prediction and ground truth, respectively. F-measure is the harmonic mean value of the precision ($prec$) and recall ($reca$), \ie, $\text{F-measure}=\frac{{(1-\beta^2)prec\cdot reca}}{\beta^2prec + reca}$, where $\beta^2$ is set to 0.3 for emphasizing the importance of precision.  E-measure captures both the local matching and region-level matching information of segmentation maps for assessment. IoU is a metric describing the extent of overlap between two segmentation maps. It is defined as $\text{IoU}=\frac{inter}{union}$, where $inter$ and $union$ indicate the intersection and union of two segmentation maps, respectively. Note that the relevant concepts of S-measure~\cite{fan2017structure} in 3D space may change, thus being ignored.

\begin{table}[t]
	\renewcommand{\tabcolsep}{1.8mm}
	\centering
	\begin{tabular}{@{}lccccc@{}}
		\toprule
		Methods & Years & MAE $\downarrow$ & F-measure $\uparrow$ & E-measure $\uparrow$ & IoU $\uparrow$ \\ \midrule
		PointNet~\cite{8099499} & \multicolumn{1}{c|}{CVPR'17} & 0.116 &0.632  &0.768  & 0.519 \\
		PointNet$++$~\cite{qi2017pointnet++} & \multicolumn{1}{c|}{NeurIPS'17} &0.077  &0.738  &0.816  &0.608  \\
		PointCNN~\cite{li2018pointcnn} & \multicolumn{1}{c|}{NeurIPS'18} &0.142  &0.409  &0.575  &0.265  \\
		ShellNet~\cite{9010996} & \multicolumn{1}{c|}{ICCV'19} &0.074  &0.753  &0.848  &0.648 \\
		RandLA~\cite{9440696} & \multicolumn{1}{c|}{TPAMI'21} &0.127  &0.633  &0.740  &0.517  \\ \midrule
		Ours & \multicolumn{1}{c|}{-} &\textbf{0.069}  &\textbf{0.769}  &\textbf{0.851}  &\textbf{0.656}  \\ \bottomrule
	\end{tabular}
	\caption{Benchmarking results of six  representative baseline models on our \textit{PCSOD} dataset. $``\uparrow"/``\downarrow"$ suggests that larger/smaller is better. Note that the best results are shown in \textbf{boldface}}
	\label{tab:1}
\end{table}

\begin{figure*}[t]
	\centering
	\includegraphics[width=0.9\linewidth]{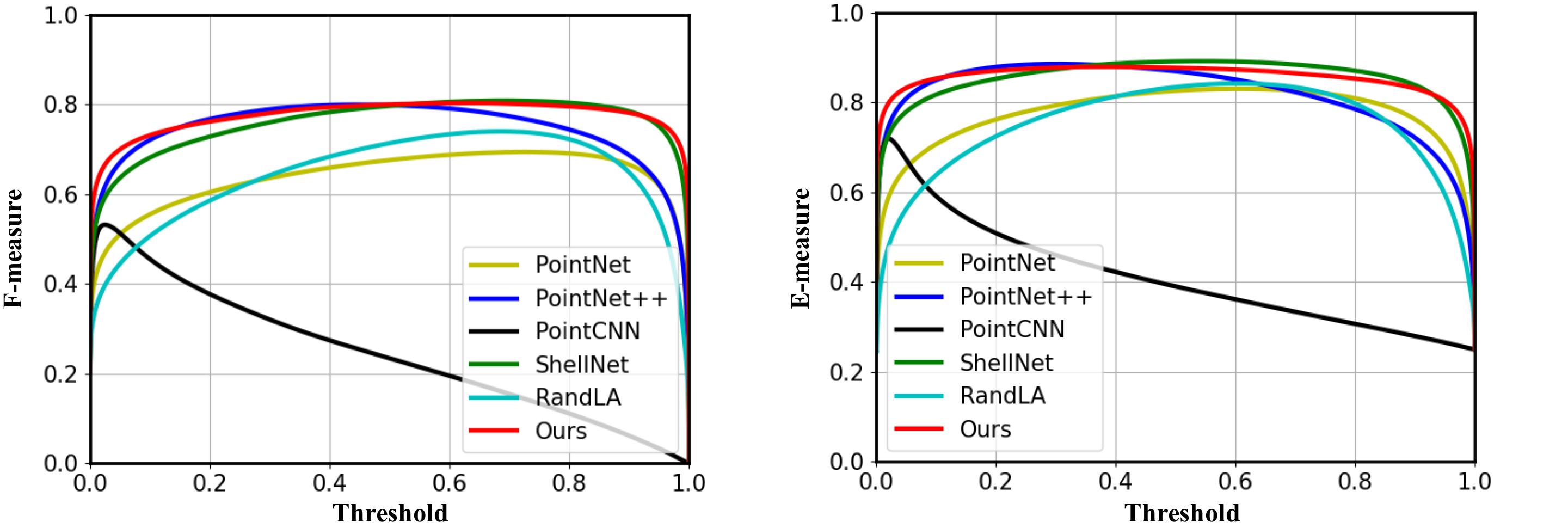}
	\caption{F-measure and E-measure under different thresholds.}
	\label{fig6}
\end{figure*}

\begin{figure*}[t]
	\centering
	\includegraphics[width=\linewidth]{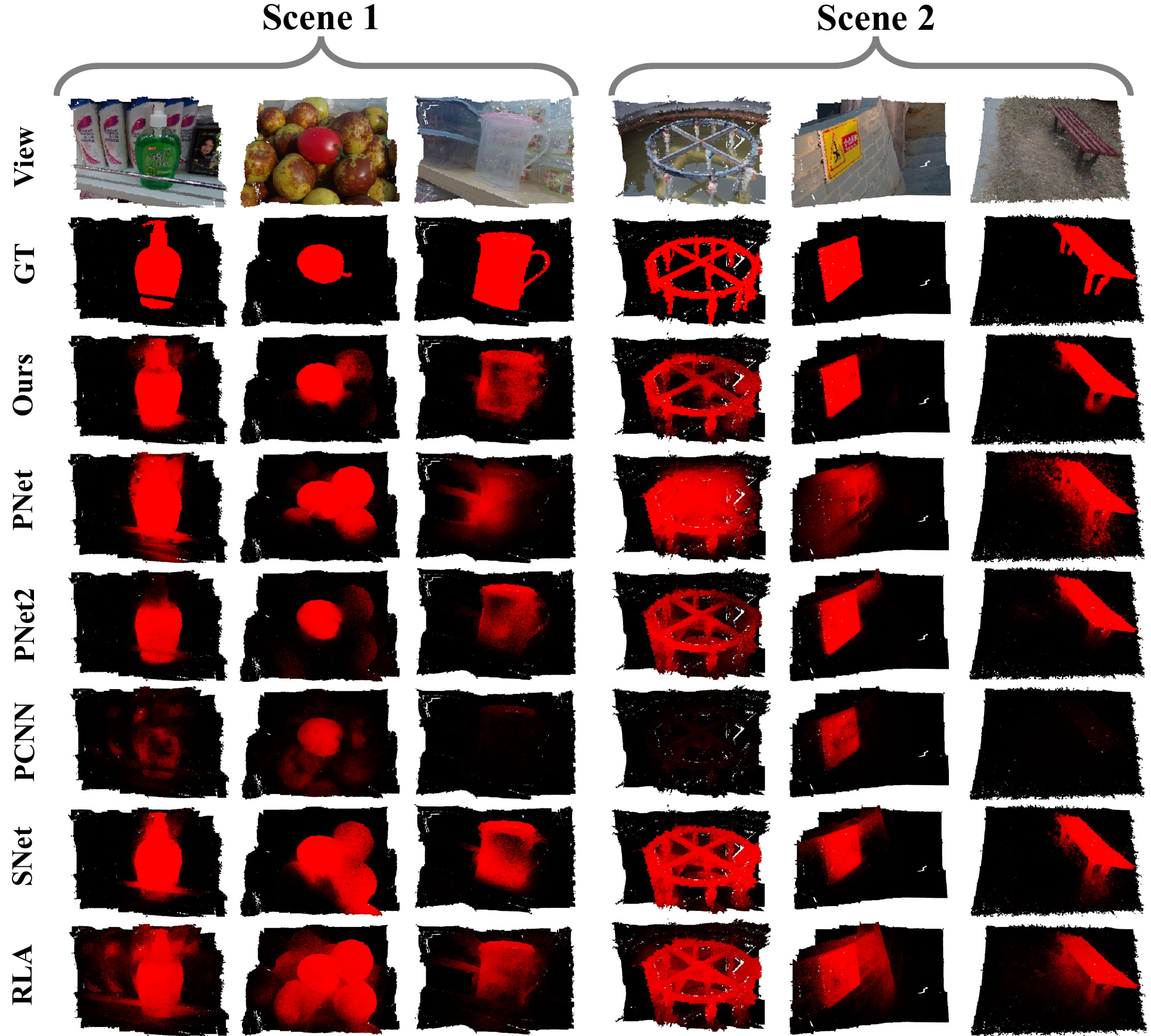}
	\caption{Qualitative comparison of six baseline models on views of two common scenes, \ie, a supermarket (Scene 1) and a park (Scene 2). Note that “GT”, “PNet”, “PNet2”, “PCNN”,  “SNet”, and “RLA” mean the ground truth, PointNet~\cite{8099499},  PointNet$++$~\cite{qi2017pointnet++}, PointCNN~\cite{li2018pointcnn}, ShellNet~\cite{9010996}, and RandLA~\cite{9440696}, respectively.}
	\label{fig7}
\end{figure*}

\subsection{Comparison and Analysis}
To the best of our knowledge, there is no deep learning-based method designed for point cloud SOD. Consequently, we introduce five representative baseline models~\cite{8099499,qi2017pointnet++,li2018pointcnn,9440696,9010996} from others segmentation tasks for comparison and analysis. PointNet~\cite{8099499} and its improved version, namely PointNet$++$~\cite{qi2017pointnet++}, are the two most representative models in point cloud processing. PointCNN~\cite{li2018pointcnn}, ShellNet~\cite{9010996}, and RandLA~\cite{9440696} indicate three promising directions of point cloud processing, \ie, powerful convolution, effective  neighborhood connection, and advanced reduction. For a fair comparison, we retrain these models on our \textit{PCSOD} dataset according to the recommended parameter settings and produce the final results by the same voting strategy as our method.   

\subsubsection{Quantitative Comparison.} In Tab.~\ref{tab:1}, we list the results of six baseline models on four evaluation metrics. We can learn that the proposed method achieves \sArt performance and outperforms all competitors by a clear margin. Specifically, our model surpasses the second-best model ShellNet by 6.8\%, 2.1\%, 0.4\%, and 1.2\% on MAE, F-measure, E-measure, and IoU. RandLA is a recently proposed model with significant superiority over PointNet$++$ for semantic segmentation. However, experiments in Tab.~\ref{tab:1} show that RandLA has no advantage on SOD and even performs worse than PointNet$++$, indicating designing tailored models for point cloud SOD is non-trivial. Though our baseline model has the best performance, there is still considerable room for performance improvement, which demands further efforts from the research community. To study the generalizability of these baseline models under different thresholds, we plot the F-measure scores and E-measure scores by taking different thresholds.  As shown in Fig.~\ref{fig6}, the results of our method are much flatter at most thresholds, which demonstrates that our method has excellent generalizability.

\subsubsection{Qualitative Comparison.} To further reveal the feasibility of our solution predicting salient objects of any given 3D views, we illustrate the results of several frequent views from two common scenes in Fig.~\ref{fig7}. Scene~1 is a supermarket (indoor scene), while Scene~2 is a park (outdoor scene), both of which are unseen by these models. It can be seen that most baseline models can locate the salient objects of given views, except for PointCNN. Though some views are very challenging, \eg, cluttered background (column 2), transparent object (column 3), complex structure (column 4),  and random view with non-central object (column 5 and 6), our method can consistently produce accurate and complete segmentation maps with high contrast, which evidences the superiority of our method.

\begin{table}[t]
	\renewcommand{\tabcolsep}{1.8mm}
	\centering
	\begin{tabular}{lccccc}
		\hline
		No. & Methods& MAE $\downarrow$ & F-measure $\uparrow$ & E-measure $\uparrow$ & IoU $\uparrow$ \\ \hline
		1 & \multicolumn{1}{c|}{PointNet$++$~\cite{qi2017pointnet++}} &0.077  &0.738  &0.816  &0.608  \\
		2 & \multicolumn{1}{c|}{+SPB} &0.076  &0.748  &0.828  & 0.624 \\ 
		3 & \multicolumn{1}{c|}{+SPB, +PPB 1} &{0.073}  &{0.754}  &{0.840}  &{0.639}  \\ 
		4 & \multicolumn{1}{c|}{+SPB, +PPB 1, +PPB 2} &\textbf{0.069}  &\textbf{0.769}  &\textbf{0.851} &\textbf{0.656}  \\\hline
		5 & \multicolumn{1}{c|}{Mean Reduction} &0.071  &0.764  &0.843  &0.649  \\
		6 & \multicolumn{1}{c|}{Max Reduction} &0.070  &{0.765}  &0.843  & 0.651 \\ 
		7 & \multicolumn{1}{c|}{Attentive Reduction~\cite{9440696}} &0.074  &{0.758}  &0.847  & \textbf{0.658} \\ 
		8 & \multicolumn{1}{c|}{ Mean-max Reduction} &\textbf{0.069}  &\textbf{0.769}  &\textbf{0.851} &{0.656} \\\hline

	\end{tabular}
	\caption{Ablation analysis of the proposed point cloud SOD model. No.1-No.4 study the effectiveness of our SPB and PPB, respectively. “PPB 1” and “PPB 2” denote the PPBs for producing global semantics and multi-scale features, respectively. No.5-No.8 investigate the alternative reduction operations.  }
	\label{tab:2}
\end{table}

\subsection{Ablation Study} To analyze the fundamentals of our baseline model, we conduct extensive ablation experiments in Tab.~\ref{tab:2}. The ablation experiments are based on the encoder PointNet$++$~\cite{qi2017pointnet++}, studying the effectiveness of the designs in our decoder, \ie, key modules and feature reduction operations. In each experiment, only one influential factor is changed as the others keep the same for a fair comparison.

\begin{figure*}[t]
	\centering
	\includegraphics[width=\linewidth]{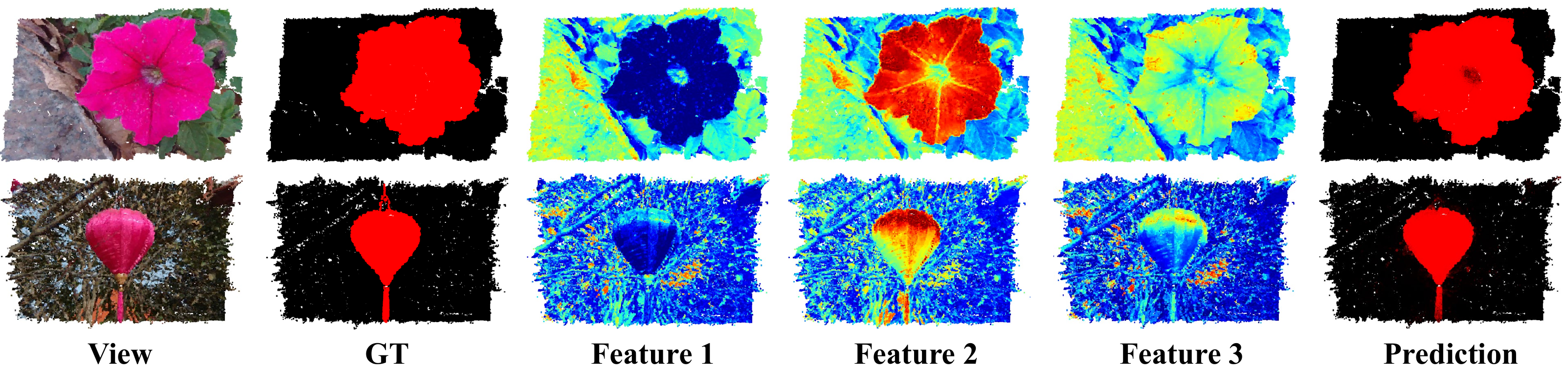}
	\caption{3D heatmap visualization of feature maps. Feature 1, Feature 2, and Feature 3 represent the multi-scale features, global semantics, and enhanced multi-scale features, respectively.}
	\label{fig8}
\end{figure*}

To investigate the contributions from our SPB and PPB separately, we first load the SPB into the encoder. By comparing No.1 and No.2 in Tab.~\ref{tab:2},  we can learn that the introduction of our SPB can help promote the performance of our model in locating salient objects. However, because the high-level features from the encoder have limited receptive fields, directly utilizing them as the semantics can only achieve suboptimal performance. As demonstrated in Tab.~\ref{tab:2} (No.3), a properly configured PPB helping acquire semantics with global receptive fields can unlock the potential of the SPB. Besides, the PPB with a different configuration can also strengthen the multi-scale representations of features, which benefits the perception of objects of different sizes. Therefore, another PPB in the ablation No.4 can bring orthogonal contributions to SOD. Fig.~\ref{fig8} further shows how the feature maps change. Due to the dilution of high-level features, multi-scale features incorrectly focus on the non-salient background, whereas the global semantics have an accurate perception of salient objects. The SPB can correct the deviation of multi-scale features by combining global semantics and obtain the enhanced multi-scale features. The ablations No.4-No.8 in Tab.~\ref{tab:2} study various reduction manners. It can be seen that our Mean-max reduction can outperform the individual Mean reduction or Max reduction. Furthermore, compared to the attentive reduction~\cite{9440696}, our method has a better performance without increasing the number of network parameters.

\section{Conclusion}

In this paper, we present the first comprehensive study on point cloud SOD, involving its formulation, dataset construction, and baseline design. To avoid the saliency conflict, we propose a novel view-dependent perspective of salient objects. Our formulation can reasonably reflect the salient objects in point cloud scenarios. Then we elaborately construct a high-quality dataset, namely \textit{PCSOD}, and contribute a baseline model for point cloud SOD. Our dataset has excellent generalizability and broad applicability, expected to boost the advance of SOD and many other vision tasks. We conduct extensive experiments on our dataset to verify the feasibility of our solution. Experimental results show that our baseline model has significant superiority and produces visually favorable predictions. Our work reveals the potential of point cloud SOD and pave the way for further study.

\textbf{Acknowledgements.} This work was supported by National Key R\&D Program of China (2020AAA0103501) , The Major Key Project of PCL, Natural Science Foundation of China (61801303, 62031013), Guangdong Basic and Applied Basic Research Foundation (2019A1515012031), Shenzhen Fundamental Research Program (GXWD20201231165807007-20200806163656003), and Shenzhen Science and Technology Plan Basic Research Project (JCYJ20190808161805519).

\bibliographystyle{splncs04}
\bibliography{egbib}
\end{document}